\documentclass[twocolumn,switch]{article}

\usepackage{preprint}
\usepackage[utf8]{inputenc}
\usepackage[T1]{fontenc}
\usepackage{cite}
\usepackage{amsmath,amssymb,amsfonts}
\usepackage{booktabs}
\usepackage{tabularx}
\usepackage{multirow}
\usepackage{siunitx}
\usepackage{graphicx}
\usepackage{textcomp}
\usepackage{xcolor}
\usepackage{microtype}
\usepackage{url}
\usepackage{hyperref}
\usepackage{float}
\usepackage{placeins}
\usepackage{dblfloatfix}
\usepackage{bm}
\usepackage{tikz}
\usetikzlibrary{calc}

\graphicspath{{figures/}}
\definecolor{citeblue}{HTML}{1796D2}

\renewcommand{\preprintstatus}{arXiv preprint, June 2026}

\AddToHook{shipout/foreground}{%
  \begin{tikzpicture}[remember picture,overlay]
    \ifnum\value{page}=1
      \node[anchor=south, text=gray, font=\footnotesize] at ([yshift=22pt]current page.south) {Corresponding author: Abdirashid Omar};
    \fi
    \ifnum\value{page}=2
      \node[anchor=south, text=gray, font=\footnotesize] at ([yshift=22pt]current page.south) {Code: \url{https://github.com/rashiedomar/Bitemporal-urban-change-Korea}};
    \fi
  \end{tikzpicture}%
}

\hypersetup{
  pdftitle={UrbanCDNet: Appearance-Robust and Boundary-Aware Bitemporal Change Detection for Korean Urban Building Monitoring},
  pdfauthor={Abdirashid Omar and Jonghyuk Park},
  pdfkeywords={change detection, remote sensing, Korean aerial imagery, building change detection, Siamese network, boundary-aware supervision},
  colorlinks=true,
  linkcolor=black,
  urlcolor=black,
  citecolor=citeblue
}

\newcommand{\urbancdnet}{UrbanCDNet}
\newcommand{\Fonec}{$\mathrm{F1}_c$}
\newcommand{\Iouc}{$\mathrm{IoU}_c$}
\newcommand{\PrecC}{$\mathrm{Prec}_c$}
\newcommand{\RecC}{$\mathrm{Rec}_c$}

\title{\urbancdnet{}: Appearance-Robust and Boundary-Aware Bitemporal Change Detection for Korean Urban Building Monitoring}

\author{
  Abdirashid Omar and Jonghyuk Park \\
  Department of Data Science, Graduate School of Kookmin University \\
  Seoul 02707, Republic of Korea
}

\begin{document}

\twocolumn[
  \begin{@twocolumnfalse}

\maketitle

\begin{abstract}
Urban building change detection from bi-temporal aerial imagery is important for redevelopment monitoring, infrastructure management, and unauthorized-construction screening, but Korean urban scenes remain difficult because changed regions are often sparse, appearance varies strongly between acquisition dates, and useful outputs must follow building footprints rather than coarse blobs.
This paper presents \urbancdnet{}, a task-specific Siamese CNN that combines appearance-robust multi-cue comparison, alignment-aware middle-scale differencing, lightweight context refinement, scene calibration, and auxiliary boundary supervision.

Experiments use a corrected AIHub-based Korean benchmark with 3,998 training, 503 validation, and 499 test pairs, and report changed-class precision, recall, F1, and IoU.
On the locked test split, \urbancdnet{} achieves 0.7335 precision, 0.7696 recall, 0.7511 F1, and 0.6014 IoU, outperforming a strong Siamese U-Net baseline (0.7108 F1, 0.5514 IoU) and the strongest external competitor, ChangeFormer-MIT-B0 (0.7107 F1, 0.5512 IoU).
Additional diagnostic slicing shows that the gain is concentrated in the operating regimes that motivated the design: on the sparse-change subset with less than 5\% changed area, F1 improves from 0.4765 to 0.6175, and on the high photometric-gap subset it improves from 0.6349 to 0.7285.
Boundary F1 at 3-pixel tolerance rises from 0.3445 to 0.4447, while object F1 at IoU 0.3 rises from 0.0690 to 0.2258.

These results indicate that, on this Korean benchmark, task-shaped temporal comparison and boundary-aware supervision matter more than generic model scale alone.

\end{abstract}

\vspace{0.35cm}

  \end{@twocolumnfalse}
]

\section{Introduction}\label{sec:introduction}
Urban building change detection from aerial image pairs is an operational remote-sensing problem with direct value for redevelopment monitoring, permit inspection, infrastructure maintenance, and rapid map updating~\cite{aisha2023,jungeun2025,kamalakar2025,seda2022,kyaw2024,r2024,zainoolabadien2020}.
In practice, however, the target is rarely a large and obvious land-cover transition.
Many Korean urban scenes contain only a small changed footprint, while the image pair may also differ in sun angle, shadow layout, exposure, and local tone.
As a result, naive temporal differencing tends to overreact to appearance shift and underperform exactly where reliable building-level change maps are most needed.

This paper studies that setting on a corrected AIHub-based Korean benchmark and reformulates a larger thesis study as a compact research preprint.
The objective is not to claim a universally best architecture, but to test a narrower proposition: can a task-specific model for sparse urban building changes outperform both a strong Siamese baseline and stronger generic attention/transformer competitors under a unified changed-class evaluation protocol?

We answer that question with \urbancdnet{}, a Siamese CNN designed around three failure modes repeatedly observed on the Korean benchmark: pseudo-change under photometric mismatch, weak recovery of sparse small edits, and loose building-footprint localization.
The model combines multi-cue temporal comparison, alignment-aware differencing at middle scales, lightweight scene-context refinement, scene calibration, and an auxiliary boundary head used only during training.

The main contributions of this work are as follows:
\begin{itemize}
  \item We present \urbancdnet{}, a task-specific bitemporal change detector for Korean urban building monitoring that explicitly targets appearance mismatch, sparse change, and footprint geometry.
  \item We evaluate the method on a corrected AIHub-based Korean benchmark with a fixed 3,998/503/499 train/validation/test split and changed-class precision, recall, F1, and IoU.
  \item We show that \urbancdnet{} improves over a strong Siamese U-Net baseline and over STANet-PAM, BIT-R18, and ChangeFormer-MIT-B0, with the clearest gains appearing in sparse-change, photometric-gap, boundary-sensitive, and false-positive-suppression diagnostics.
\end{itemize}

\section{Related Work}\label{sec:related_work}
Recent surveys show that remote-sensing change detection has shifted from simple differencing toward deep Siamese encoders, attention-based fusion, transformers, and increasingly task-specific hybrid designs~\cite{devansh2025,aleissaee2023transformersrs,ma2019deeplearningrs,ball2017survey,ghamisi2019fusion,chen2025imbalance}.
Shared-weight Siamese encoders remain a strong starting point because they keep the two times in a common feature space and make temporal comparison well posed~\cite{ronneberger2015unet,daudt2018fully}.

CNN-based change detectors have explored denser skip fusion, residual refinement, and semantic or spatially guided comparison.
Representative examples include early multiscale Siamese CNNs, SNUNet-CD, deeply supervised fusion, TinyCD, LightCDNet, DASNet, asymmetric Siamese semantic change detectors, adaptive Siamese fusion, and enhanced-change multitask designs~\cite{hongruixuan2019,fang2021snunetcd,zhang2020deeply,codegoni2022tinycd,10214556,jie2020,kunping2022,yunzuo2025,xibing2025}.
These methods are efficient and often strong, but plain differencing remains vulnerable when changed buildings are sparse and appearance mismatch dominates the scene.

Attention-based models try to improve that comparison stage by reweighting useful spatial or channel cues.
STANet remains a standard reference, while HANet, Change Guiding Network, SemiSANet, AFDE-Net, ADDEDNet, three-branch attention fusion, MDANet, M2F2Net, and flow-guided semantic change models extend this direction in different ways~\cite{chen2020stanet,10093022,10234560,chengzhen2022,s2023,junheng2025,yan2024,shanshan2024,binhao2025,k2024}.
These methods show that better fusion helps, but they do not necessarily solve the specific combination of sparse change, appearance stress, and tight footprint localization required in Korean urban building monitoring.

Transformer-based approaches strengthen temporal interaction even further.
BIT and ChangeFormer are strong generic baselines, and later variants such as STransUNet, multitask CNN--Transformer models, STeInFormer, Changer, foundation-model-based CD, and newer benchmark-driven systems continue to improve global reasoning~\cite{chen2022bit,bandara2022changeformer,jianghua2022,wei2024,xiaowen2024,fang2022changer,chen2023time,10438490,liu2025jl1}.
Open-CD has also made cross-model comparison easier by standardizing training and evaluation pipelines for many of these architectures~\cite{li2024opencd}.

For building-focused change detection, boundary quality matters as much as aggregate region overlap.
S2Looking and AERNet emphasize that high-resolution building CD is not only a binary segmentation problem but also a footprint-localization problem~\cite{shen2021s2looking,zhang2023aernet}.
This motivates explicit edge-aware supervision, which has also been effective in broader dense prediction settings~\cite{yu2017casenet,sudre2017dice}.
UrbanCDNet is positioned in this narrower space: rather than claiming universal architectural superiority, it asks whether a model explicitly shaped around appearance robustness, sparse-change recovery, and boundary quality can win on the corrected Korean AIHub benchmark~\cite{nia2022aihub}.

\section{Method}\label{sec:method}
\begin{figure*}[!t]
  \centering
  \begin{tikzpicture}
    \node[inner sep=0] (img) {\includegraphics[width=\textwidth,trim=0 0 0 72,clip]{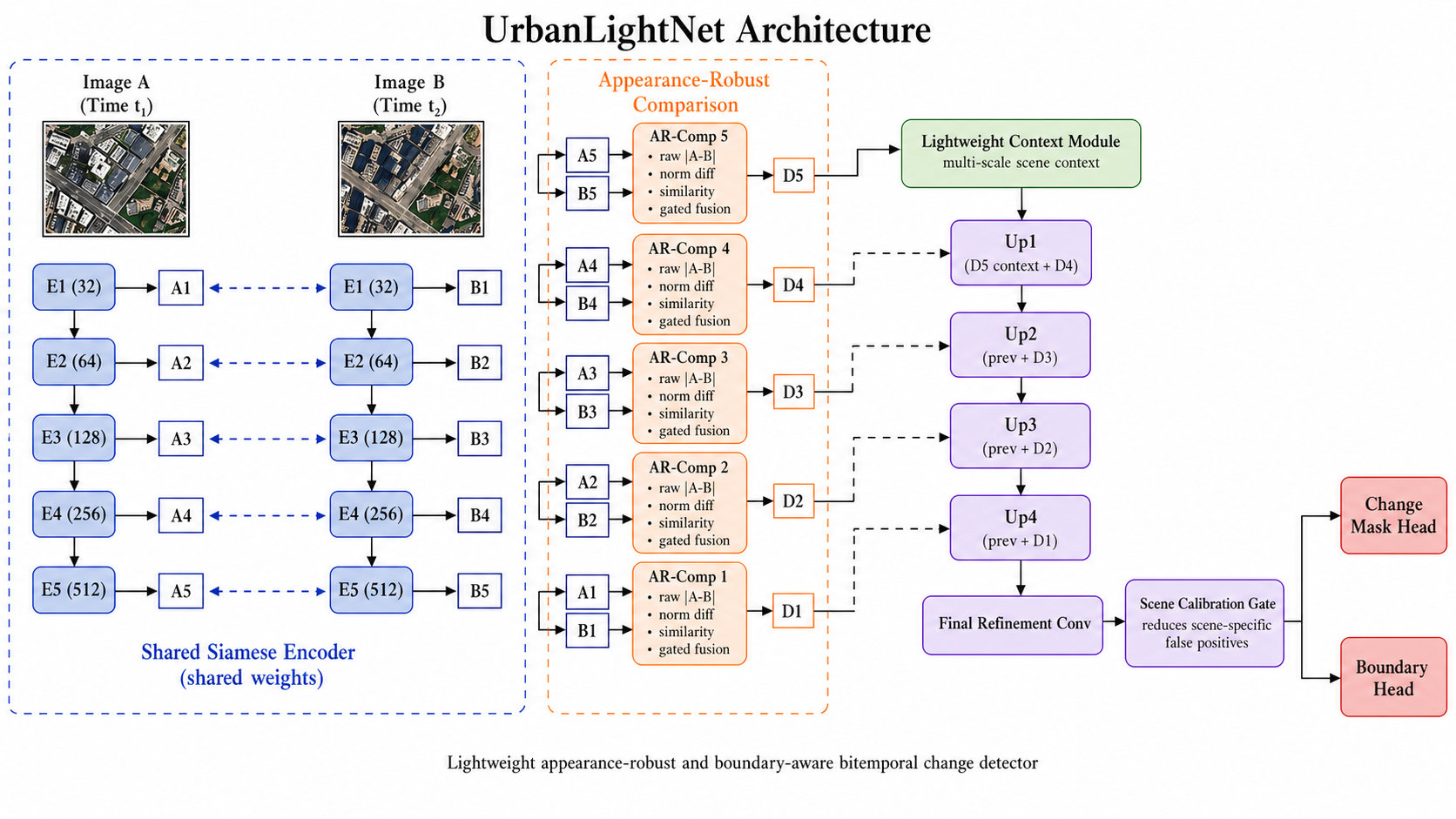}};
    \node[fill=white, fill opacity=0.95, text opacity=1, minimum width=0.48\textwidth, minimum height=0.45cm, rounded corners=2pt] at ([yshift=0.55cm]img.south) {Lightweight appearance-robust and boundary-aware bitemporal change detector};
  \end{tikzpicture}
  \caption{UrbanCDNet overview. A shared Siamese encoder extracts multiscale features from the bi-temporal pair. The model then applies appearance-robust comparison, alignment-aware middle-scale matching, lightweight context refinement, scene calibration, and an auxiliary boundary head.}
  \label{fig:arch}
\end{figure*}

UrbanCDNet is a Siamese CNN-based bitemporal change detector organized around three ideas: appearance-robust temporal comparison, lightweight scene-aware refinement, and boundary-aware training.
Figure~\ref{fig:arch} summarizes the full pipeline.

\subsection{Appearance-Robust Comparison}
Let $I_a$ and $I_b$ be the two RGB inputs.
A shared encoder produces multiscale features $F_a^{(s)}, F_b^{(s)} \in \mathbb{R}^{C_s \times H_s \times W_s}$.
For each scale, UrbanCDNet does not rely on raw absolute difference alone.
Instead, it combines three complementary cues:
\begin{equation}
  D_{\mathrm{raw}}^{(s)} = \left|F_a^{(s)} - F_b^{(s)}\right|,\quad
  D_{\mathrm{norm}}^{(s)} = \left|\bar{F}_a^{(s)} - \bar{F}_b^{(s)}\right|,\quad
  S^{(s)} = \bar{F}_a^{(s)} \odot \bar{F}_b^{(s)},
  \label{eq:comparison_cues}
\end{equation}
where $\bar{F}$ denotes instance-normalized and channelwise $\ell_2$-normalized features.
The normalized cues reduce sensitivity to photometric shift, while the similarity term retains local temporal agreement.

A learned gate then fuses the cues into a refined comparison tensor:
\begin{equation}
  G^{(s)} = \sigma\!\left(\phi_g^{(s)}\!\left([D_{\mathrm{raw}}^{(s)}; D_{\mathrm{norm}}^{(s)}; S^{(s)}]\right)\right),
  \label{eq:gate}
\end{equation}
\begin{equation}
  \hat{D}^{(s)} = \phi_r^{(s)}\!\left([(1-G^{(s)}) \odot D_{\mathrm{raw}}^{(s)};\, G^{(s)} \odot D_{\mathrm{norm}}^{(s)};\, S^{(s)}]\right),
  \label{eq:refined_compare}
\end{equation}
where $\phi_g^{(s)}$ and $\phi_r^{(s)}$ are learned $1\times1$ fusion blocks.
This keeps strong raw-difference evidence when structural change is clear, but shifts the representation toward normalized comparison when the pair is dominated by illumination or shadow differences.

\subsection{Alignment, Context, and Scene Calibration}
Middle scales are additionally given a locally aligned cue to reduce residual spatial mismatch near building edges:
\begin{equation}
  \bar{D}_{\mathrm{align}}^{(s)}
  =
  D_{\mathrm{raw}}^{(s)} + \gamma^{(s)}
  \left(\left|F_a^{(s)} - W(F_b^{(s)}, \delta^{(s)})\right| - D_{\mathrm{raw}}^{(s)}\right),
  \quad s \in \{3,4\},
  \label{eq:alignment}
\end{equation}
where $W(\cdot,\delta^{(s)})$ denotes bilinear warping under a learned local displacement field and $\gamma^{(s)}$ is a residual scale initialized near zero.
The aligned difference is injected through the same gated comparison path rather than replacing the raw cue outright.

At the deepest scale, UrbanCDNet applies a lightweight ASPP-style context module to enlarge the receptive field before decoding.
After the decoder, a residual channel gate driven by global average pooling recalibrates the final feature map to suppress scene-specific false positives that survive local comparison.

\subsection{Boundary-Aware Training}
The final decoder feature produces both a change-mask head and an auxiliary boundary head.
The boundary branch is used only during training, where it encourages tighter footprint geometry.
The overall loss is
\begin{equation}
  \mathcal{L}
  =
  \mathcal{L}_{\mathrm{BCE}}
  + 0.5\,\mathcal{L}_{\mathrm{Dice}}
  + 0.05\,\mathcal{L}_{\mathrm{Bnd}},
  \label{eq:loss}
\end{equation}
where $\mathcal{L}_{\mathrm{Dice}}$ follows the standard overlap formulation for imbalanced segmentation~\cite{sudre2017dice}.
This design keeps inference simple while making boundary quality an explicit optimization target.

\section{Experiments}\label{sec:experiments}
Experiments use a corrected raw-pair, location-safe Korean benchmark derived from the AIHub urban building change dataset~\cite{nia2022aihub}.
The fixed split contains 3,998 training, 503 validation, and 499 test pairs.
The benchmark is strongly imbalanced: the mean changed-pixel ratio of the test split is 0.0706, the median is 0.0492, and 253 of 499 test pairs (50.7\%) contain less than 5\% changed area.
This is why we report changed-class precision, recall, F1, and IoU rather than mean-over-classes summaries.
Thresholds are selected on validation and then applied once to the locked test split.

The compared models are a strong Siamese U-Net baseline, STANet-PAM~\cite{chen2020stanet}, BIT-R18~\cite{chen2022bit}, ChangeFormer-MIT-B0~\cite{bandara2022changeformer}, and \urbancdnet{}.
External competitors are trained through Open-CD~\cite{li2024opencd} and evaluated through the same changed-class code path.
All models use a 512 crop during training except STANet-PAM, which is restricted to 256 for memory reasons.

\subsection{Overall Performance}
Table~\ref{tab:main_results} shows that \urbancdnet{} ranks first on both \Fonec{} and \Iouc{}.
The strongest external competitor is ChangeFormer-MIT-B0, which attains slightly higher precision but clearly lower recall.
UrbanCDNet therefore wins not by overpredicting easy positives, but by reaching a better precision--recall balance on the changed-building class.
\begin{table*}[!t]
  \centering
  \caption{Changed-class test performance on the corrected Korean AIHub benchmark. Validation thresholds are selected once and then applied to the locked test split.}
  \label{tab:main_results}
  \small
  \setlength{\tabcolsep}{5pt}
  \begin{tabular*}{\textwidth}{@{\extracolsep{\fill}} lcccccc}
    \toprule
    Model & Params (M) & \PrecC{} & \RecC{} & \Fonec{} & \Iouc{} & Best val. thr. \\
    \midrule
    Baseline Siamese U-Net & 33.23 & 0.7172 & 0.7045 & 0.7108 & 0.5514 & 0.80 \\
    STANet-PAM & 13.36 & 0.5136 & 0.6795 & 0.5850 & 0.4134 & 0.90 \\
    BIT-R18 & 2.99 & 0.6677 & 0.6015 & 0.6328 & 0.4629 & 0.30 \\
    ChangeFormer-MIT-B0 & 3.85 & 0.7415 & 0.6824 & 0.7107 & 0.5512 & 0.30 \\
    \textbf{UrbanCDNet (ours)} & \textbf{22.35} & \textbf{0.7335} & \textbf{0.7696} & \textbf{0.7511} & \textbf{0.6014} & \textbf{0.75} \\
    \bottomrule
  \end{tabular*}
\end{table*}

\subsection{Small-Change and Photometric-Gap Results}
The hardest part of the benchmark is not the dense-change subset but the regime in which changed buildings occupy only a small fraction of the image.
Table~\ref{tab:change_bins} shows that UrbanCDNet leads across all change-density bins and yields its clearest gain in the less-than-5\% subset.
\begin{table}[H]
  \centering
  \caption{Changed-class F1 by change-density bin.}
  \label{tab:change_bins}
  \small
  \setlength{\tabcolsep}{3pt}
  \begin{tabular*}{\columnwidth}{@{\extracolsep{\fill}} lcccc}
    \toprule
    Model & $<5\%$ & 5--10\% & 10--20\% & $\geq$20\% \\
    \midrule
    Baseline & 0.4765 & 0.6657 & 0.7056 & 0.7654 \\
    STANet & 0.4333 & 0.5834 & 0.6680 & 0.7086 \\
    BIT & 0.5236 & 0.6216 & 0.6947 & 0.6844 \\
    ChangeFormer & 0.6025 & 0.7096 & 0.7386 & 0.8000 \\
    \textbf{UrbanCDNet} & \textbf{0.6175} & \textbf{0.7395} & \textbf{0.7759} & \textbf{0.8119} \\
    \bottomrule
  \end{tabular*}
\end{table}

This matters because the sparse subset is not a corner case.
It covers more than half of the test set and is the operating regime where false expansion and missed small edits are most damaging.
Relative to the baseline, \urbancdnet{} improves F1 from 0.4765 to 0.6175 in that subset, while still remaining ahead of ChangeFormer.

Appearance stress shows the same pattern.
Table~\ref{tab:gap_bins} reports low-, medium-, and high-gap bins using the same test set.
UrbanCDNet improves over the baseline in every bin and remains slightly ahead of ChangeFormer under the high-gap condition.
\begin{table}[H]
  \centering
  \caption{Changed-class F1 by photometric-gap bin.}
  \label{tab:gap_bins}
  \small
  \setlength{\tabcolsep}{4pt}
  \begin{tabular*}{\columnwidth}{@{\extracolsep{\fill}} lccc}
    \toprule
    Model & Low & Medium & High \\
    \midrule
    Baseline & 0.6560 & 0.6369 & 0.6349 \\
    STANet & 0.5956 & 0.5587 & 0.5947 \\
    BIT & 0.6280 & 0.6180 & 0.6468 \\
    ChangeFormer & 0.7055 & 0.6948 & 0.7257 \\
    \textbf{UrbanCDNet} & \textbf{0.7432} & \textbf{0.7260} & \textbf{0.7285} \\
    \bottomrule
  \end{tabular*}
\end{table}

These two tables together support the same interpretation: the model gain is not confined to average-case scenes, but persists exactly where a robust building-change detector should help most.
The two slices are also complementary rather than redundant.
The change-density table stresses geometric scale, while the photometric-gap table stresses nuisance variation that does not correspond to real urban development.
UrbanCDNet remains ahead in both views, which suggests that the method is not simply tuned for one narrow artifact of the benchmark.
Instead, the normalized comparison cues and scene calibration appear to improve the model's ability to reject non-structural temporal differences without sacrificing sensitivity to small true building edits.

\subsection{Boundary and Object Localization}
Pixel F1 alone does not fully describe building-change quality, because loose masks can still score reasonably under area overlap.
Table~\ref{tab:boundary_object} therefore reports boundary-sensitive and object-level diagnostics for the baseline and UrbanCDNet.
\begin{table*}[!t]
  \centering
  \caption{Boundary-sensitive and object-level comparison between the baseline and UrbanCDNet.}
  \label{tab:boundary_object}
  \scriptsize
  \setlength{\tabcolsep}{3pt}
  \resizebox{\textwidth}{!}{%
  \begin{tabular}{lcccccc}
    \toprule
    Model & Boundary F1@2px & Boundary F1@3px & Boundary F1@3px ($<5\%$) & Object F1@0.1 & Object F1@0.3 & Object F1@0.5 \\
    \midrule
    Baseline Siamese U-Net & 0.2894 & 0.3445 & 0.3022 & 0.0767 & 0.0690 & 0.0603 \\
    \textbf{UrbanCDNet} & \textbf{0.3818} & \textbf{0.4447} & \textbf{0.4279} & \textbf{0.2392} & \textbf{0.2258} & \textbf{0.2021} \\
    \bottomrule
  \end{tabular}
  }
\end{table*}

Boundary F1 improves by about ten points at 3-pixel tolerance, and the gain is larger still on the sparse-change subset.
Object-level F1 remains numerically harsh for both models, but the relative gain is large enough to show that UrbanCDNet is not merely expanding changed regions; it is recovering more coherent building instances.
The sparse-change mean absolute area-ratio error also drops from 0.0448 to 0.0260, a 42\% reduction.
This distinction matters for municipal monitoring workflows.
In many practical review settings, a detector is used to trigger manual inspection at the building footprint level rather than to estimate only total changed area.
A mask that is roughly in the right place but bleeds across roofs, roads, or cast shadows still increases downstream verification cost.
The boundary and object metrics therefore strengthen the claim that UrbanCDNet produces outputs that are more usable for building-centered screening, not only better under aggregate pixel overlap.
The object-level scores are especially informative because they penalize two common urban failure modes at once.
First, if a model breaks one changed building into several small fragments, the object match becomes unstable even when some changed pixels are correctly activated.
Second, if a model merges the target building with nearby clutter such as roads, parking lots, or neighboring roofs, the resulting mask may still retain moderate pixel overlap while failing as an actionable building instance.
UrbanCDNet improves under both pressures because the gain is not limited to looser overlap thresholds but continues through the stricter object settings.

Another useful way to read Table~\ref{tab:boundary_object} is as evidence about error shape rather than only error amount.
The baseline already captures part of the changed area in many samples, which is why its pixel F1 is not poor in absolute terms.
However, the boundary and object diagnostics reveal that many of those detections remain geometrically untidy.
UrbanCDNet narrows that gap by producing masks whose support is better matched to the real footprint.
This is consistent with the training design: the auxiliary boundary branch does not try to invent new semantic evidence, but it regularizes how the evidence is expressed spatially.

That geometric effect is also important for the sparse-change subset, where changed buildings occupy only a very small portion of the image.
In that regime, a few extra pixels around the footprint can double the apparent object area and quickly turn a useful detection into a noisy alert.
The large improvement in boundary F1 on the less-than-5\% subset therefore indicates that UrbanCDNet is not simply winning on large easy changes elsewhere in the benchmark.
It is improving precisely where footprint discipline is most fragile and where an operational reviewer would be least tolerant of spatial spillover.

\subsection{Qualitative Results}
\begin{figure*}[!t]
  \centering
  \includegraphics[width=0.985\textwidth]{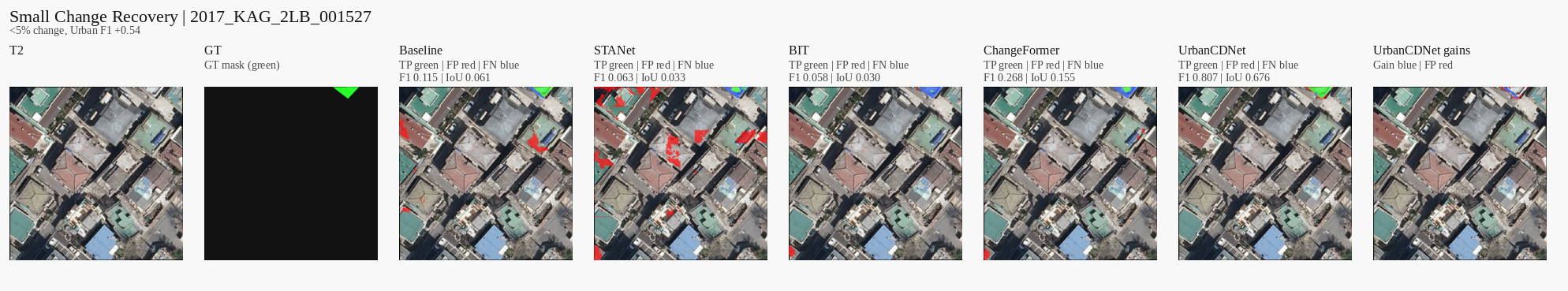}\\[-0.2em]
  \includegraphics[width=0.985\textwidth]{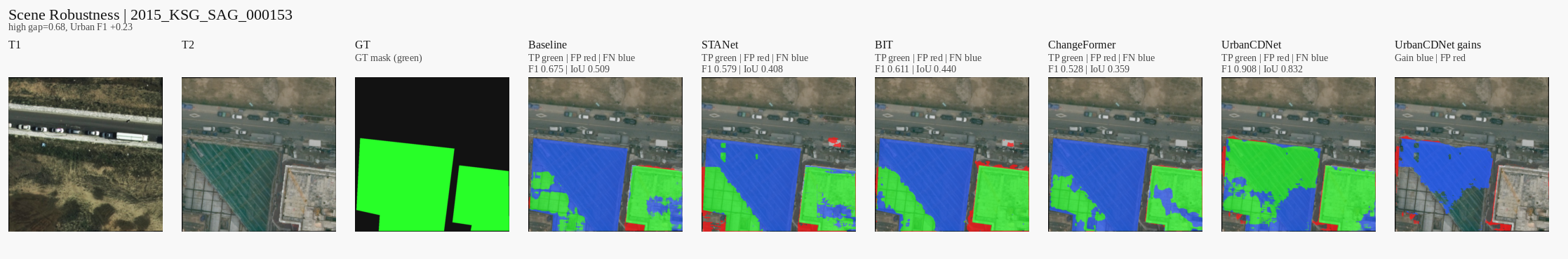}\\[-0.2em]
  \includegraphics[width=0.985\textwidth]{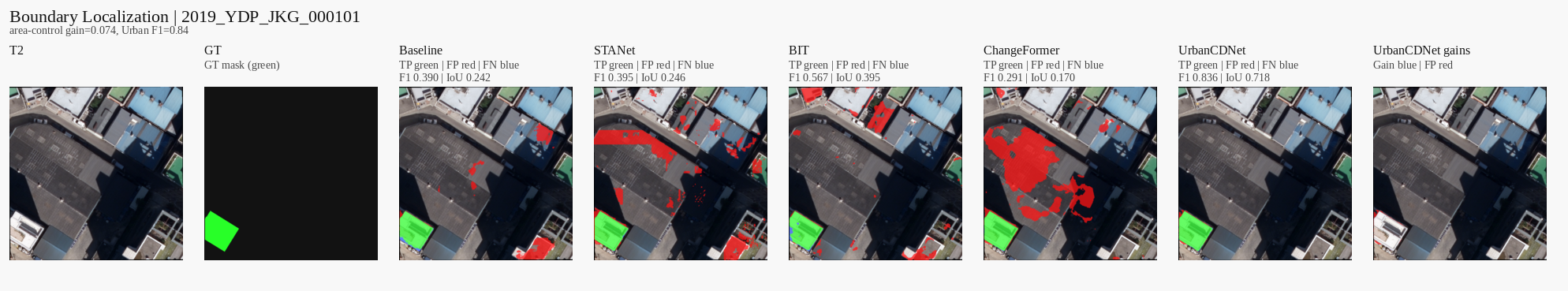}\\[-0.2em]
  \includegraphics[width=0.985\textwidth]{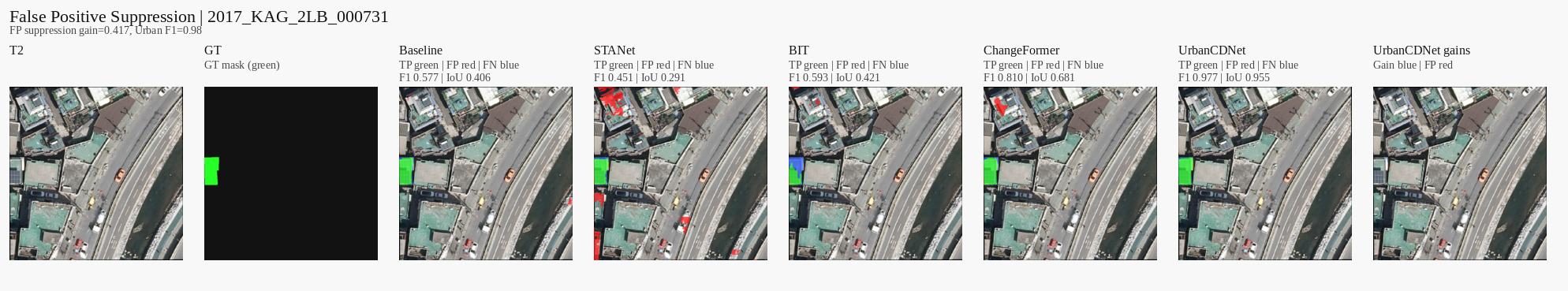}
  \caption{One representative qualitative example per error category. From top to bottom: sparse small-change recovery, appearance-gap robustness, boundary localization, and false-positive suppression. UrbanCDNet is consistently more selective under appearance stress while preserving the target footprint more tightly.}
  \label{fig:qualitative_page}
\end{figure*}

Figure~\ref{fig:qualitative_page} provides one representative example for each of the four recurring qualitative stories observed throughout the thesis-scale study: sparse small-change recovery, appearance robustness, tighter boundary localization, and false-positive suppression.
Each row uses the same comparison layout across Baseline, STANet, BIT, ChangeFormer, and UrbanCDNet.
These examples are not meant as isolated visual anecdotes.
They were selected because each row matches a failure mode that also appears in the quantitative diagnostics above, allowing the reader to connect the numerical gains to concrete prediction behavior.
In the sparse-change case, the main difference is recall on small edited rooftops.
In the appearance-gap and false-positive cases, the key effect is better selectivity under shadows, seasonal tone shift, and local illumination mismatch.
In the boundary row, the most visible gain is that UrbanCDNet tracks the building footprint more closely instead of filling large amorphous regions around the target.

These representative rows are consistent with the quantitative tables.
UrbanCDNet misses fewer sparse building changes than the baseline, avoids many pseudo-change regions produced under strong photometric mismatch, localizes the footprint more tightly, and suppresses large false-positive blobs that remain in the generic competitors.
Just as importantly, the four rows show that the same model behavior repeats across qualitatively different scenes rather than only within one easy visual pattern.
That consistency is useful on the Korean benchmark because the dominant source of error changes from sample to sample: some pairs are difficult because the changed building is tiny, others because the radiometric gap is large, and others because boundaries are cluttered by dense urban context.
UrbanCDNet does not eliminate every failure, but it shifts the error profile toward smaller and more localized mistakes, which is the direction that matters most for a practical pre-screening system.
\FloatBarrier

\subsection{Component Ablation}
The component study in Table~\ref{tab:ablation} evaluates whether UrbanCDNet's gains come from one dominant block or from the interaction of the full design.
All removals hurt performance, and the largest drop appears when the boundary head is removed.
This is consistent with the benchmark itself: the labels are building footprints, so geometry-sensitive supervision is not optional detail but a core part of the task.
Removing the context module or scene calibration also causes clear losses, which supports the claim that many benchmark errors are scene-level false positives rather than purely local mistakes.
The alignment block contributes a smaller but still meaningful improvement.

\begin{table}[H]
  \centering
  \caption{Ablation of key UrbanCDNet components on the Korean benchmark.}
  \label{tab:ablation}
  \small
  \setlength{\tabcolsep}{4pt}
  \begin{tabular*}{\columnwidth}{@{\extracolsep{\fill}} lccc}
    \toprule
    Variant & \Fonec{} & \Iouc{} & $\Delta$\Fonec{} \\
    \midrule
    UrbanCDNet & 0.7511 & 0.6014 & 0.0000 \\
    w/o alignment & 0.7234 & 0.5667 & -0.0277 \\
    w/o context module & 0.7138 & 0.5549 & -0.0373 \\
    w/o scene calibration & 0.7202 & 0.5627 & -0.0309 \\
    w/o boundary head & 0.7030 & 0.5421 & -0.0481 \\
    \bottomrule
  \end{tabular*}
\end{table}

These ablations should be read as directional evidence rather than a bit-for-bit replay of the locked winning run, but they remain useful for ranking component importance.
Taken together with the main comparison, they suggest that UrbanCDNet wins because it combines appearance-robust temporal comparison with explicit control over footprint geometry and scene-specific bias.

\FloatBarrier

\section{Conclusion}\label{sec:conclusion}
This paper presented \urbancdnet{}, a short-form research preprint distilled from a larger thesis study on Korean urban building change detection.
The method combines multi-cue temporal comparison, alignment-aware differencing, lightweight context refinement, scene calibration, and boundary-aware training.
On the corrected AIHub-based Korean benchmark, UrbanCDNet achieves the best overall changed-class performance among the tested models, improving the locked test result to 0.7511 F1 and 0.6014 IoU.

The main value of the model is not only the aggregate score increase.
Its strongest gains appear in the difficult regimes that matter operationally: sparse small changes, large appearance gap, and footprint-level localization.
Under those conditions, a task-shaped CNN remains competitive with and, on this benchmark, stronger than more generic attention and transformer baselines.

\FloatBarrier

\begingroup
\small
\bibliographystyle{unsrt}
\bibliography{refs}
\endgroup

\end{document}